\documentclass[letterpaper, 10 pt, journal, twoside]{IEEEtran}  

\IEEEoverridecommandlockouts                              

\usepackage{multicol}
\usepackage{multirow}
\usepackage{color}
\usepackage{xcolor}
\usepackage{graphicx}
\usepackage{bm}
\usepackage{amsmath}
\usepackage{amssymb}
\usepackage{tabularx}
\usepackage{subcaption}
\usepackage{url}
\usepackage{cite}
\usepackage{prettyref}
\usepackage{booktabs}
\usepackage{siunitx}

\usepackage{colortbl}
\newrefformat{fig}{Figure~\ref{#1}}
\newrefformat{sec}{Section~\ref{#1}}
\newrefformat{tab}{Table~\ref{#1}}
\newrefformat{alg}{Algorithm~\ref{#1}}
\newrefformat{eqn}{Equation~\ref{#1}}

\usepackage[symbol]{footmisc}

\usepackage{hyperref}

\makeatletter
\newcommand{\printfnsymbol}[1]{%
  \textsuperscript{\@fnsymbol{#1}}%
}
\g@addto@macro{\UrlBreaks}{\UrlOrds}
\makeatother

\title{
Autonomous Social Distancing in Urban Environments using a Quadruped Robot  
} 

\author{Tingxiang Fan$^{1*}$, Zhiming Chen$^{2*}$, Xuan Zhao$^{3}$, Jing Liang$^{4}$, Cong Shen$^{5}$, Dinesh Manocha$^{4}$, Jia Pan$^{1 \dag}$ and Wei Zhang$^{2 \dag}$
\thanks{* denotes equal contribution. $\dag$ denotes the corresponding author.}
\thanks{$^{1}$Tingxiang Fan, Jia Pan (email: {\tt\small jpan@cs.hku.hk}) are with the Department of Computer Science, The University of Hong Kong, Hong Kong, China}
\thanks{$^{2}$Zhiming Chen, Wei Zhang (email: {\tt\small zhangw3@sustech.edu.cn}) are with the Department of Mechanical \& Energy Engineering, Southern University of Science and Technology, Shenzhen, Guangdong, China}
\thanks{$^{3}$Xuan Zhao is with the Department of Biomedical Engineering, City University of Hong Kong, Hong Kong, China}
\thanks{$^{4}$Dinesh Manocha, Jing Liang are with Department of Computer Science, University of Maryland, College Park, USA}
\thanks{$^{5}$Cong Shen is with  School of Mechanical Science \& Engineering, Huazhong University of Science and Technology, Wuhan, Hubei, China}
}

\begin{document}
\maketitle

\begin{abstract}
COVID-19 pandemic has become a global challenge faced by people all over the world. Social distancing has been proved to be an effective practice to reduce the spread of COVID-19. Against this backdrop, we propose that the surveillance robots can not only monitor but also promote social distancing. Robots can be flexibly deployed and they can take precautionary actions to remind people of practicing social distancing. In this paper, we introduce a {\em fully autonomous} surveillance robot based on a quadruped platform that can promote social distancing in complex urban environments. Specifically, to achieve autonomy, we mount multiple cameras and a 3D LiDAR on the legged robot. The robot then uses an onboard real-time social distancing detection system to track nearby pedestrian groups. Next, the robot uses a crowd-aware navigation algorithm to move freely in highly dynamic scenarios. The robot finally uses a crowd-aware routing algorithm to effectively promote social distancing by using human-friendly verbal cues to send suggestions to over-crowded pedestrians. We demonstrate and validate that our robot can be operated autonomously by conducting several experiments in various urban scenarios.

\end{abstract}


\section{Introduction}
\label{sec:intro}

\IEEEPARstart{C} OVID-19 pandemic has quickly become the most dramatic and disruptive event experienced by people all over the world. People may need to live with the virus for a long time. One of the most effective measures to minimize the spread of the coronavirus is to promote social distancing. To achieve it, some related applications are developed in the existing on-site closed-circuit television (CCTV) systems to detect social distancing. However, the on-site monitoring system is not ubiquitous in some areas and sometimes it may not be able to cover all public corners. Furthermore, although this sort of monitoring system has detected social distancing violations, it fails to take any proactive actions to promote social distancing.   

Compared to the on-site monitoring system, the surveillance robots can be flexibly deployed and patrol in the desired public areas. Moreover, the robot can take precautions to promote social distancing rather than monitoring them. These potential applications have been validated by teleoperation robots\cite{sg_dog_robot}  and hybrid systems\cite{COVID-ROBOT-DM}. The hybrid system introduces the external devices such as CCTV to help robots monitoring social distancing. However, developing such a fully autonomous surveillance robot in complex urban environments without any external device still encounter several challenges. First, to monitor the social distance between pedestrians without any external device, a robot-centric real-time perception system is demanded on the on-board devices with limited computation. Second, in many urban scenarios, the robot needs to safely navigate through unstructured and highly dynamic environments. Third, more intelligent interactions with humans need to be designed to improve the efficiency of promoting social distancing. 

In this paper, we introduce a fully autonomous surveillance robot to promote social distancing in complex urban environments. To achieve this autonomy, we first build the surveillance system with multiple cameras and a 3D LiDAR on a legged robot, which empowers the robot omni-perceptibility and extends its traversability in complex urban terrains with uneven terrains and stairs, which are challenging for normal wheeled mobile robot. Then, we develop an on-board real-time social distancing detection system with the ability to track the robot's nearby pedestrian groups. Next, the CrowdMove~\cite{fan2018crowdmove} algorithm is used to navigate the robot in highly dynamic environments. Finally, we develop a 
crowd-aware routing algorithm to allow robots to approach over-crowded pedestrian groups and effectively promote social distancing using verbal cues. We also investigate the influence of human voices to the effectiveness and acceptability of quadruped surveillance and social distancing, because it has been reported that a robotic patrolling inspector can be terrifying for general citizen\footnote[7]{\url{https://www.fastcompany.com/90539438/this-covid-swabbing-robot-is-terrifying-but-it-doesnt-need-to-be}}.
We demonstrate that this surveillance robot can be automatically operated with satisfactory human response by conducting experiments in various urban scenarios.

The rest of this paper is organized as follows. Section~\ref{sec:related} reviews the related work. Section~\ref{sec:hardware} describes the hardware platform that the surveillance robot builds on. Section~\ref{sec:tracking} presents the robot's tracking algorithm used for social distancing detection. Section~\ref{sec:nav} illustrates the robot's navigation in urban scenarios. Section~\ref{sec:interaction} discusses the robot's interactions with humans through verbal 
communication. Section~\ref{sec:exp} presents the experiments conducted to validate the proposed algorithms. Section~\ref{sec:conclusion} concludes this paper.

\section{Related Work}
\label{sec:related}
In this section, we will give a brief overview of algorithms related to our system, including the robotic perception, navigation, and interaction for surveillance robots.

\subsection{Perception for Surveillance Systems}
Pedestrian tracking has been widely applied in surveillance video analysis and is well developed based on research on multi-object tracking problems~\cite{chandra2019densepeds,chandra2019robusttp,fang2018recurrent,wojke2017simple}.

Discrete velocities are used to model pedestrians' motion~\cite{antonini2006behavioral,robin2009specification}. Although discretization improves the efficiency of prediction, this approach cannot fully satisfy real-life continuous situations. Chung et al. developed cognitive models to improve the performance of their model~\cite{chung2010mobile}, but they did not consider the people's facing direction by only using circulars to model pedestrians.

Helbing et al. proposed social force to model and predict people's move according to energy potential which is caused by people and obstacles~\cite{helbing1995social}. Then the tracking performance is improved by detecting abnormal events among pedestrians~\cite{mehran2009abnormal}. Pellegrini et al. developed Linear Trajectory Avoidance (LTA) to improve the accuracy of motion prediction~\cite{pellegrini2009you}. \cite{pellegrini2010improving, choi2009they, yamaguchi2011you} developed social interactions among pedestrians to improve the accuracy of behavior models. Sheng et al. proposed the Robust Local Effective Matching Model to solve the issue of partial detection of objects~\cite{sheng2017robust}. However, these approaches cannot describe pedestrians' dynamics in dense situations because they only use linear models. In our system, a nonlinear model, Frontal RVO (F-RVO)~\cite{chandra2019densepeds} is used to simulate motions in crowds and also model the dynamic behaviors considering pedestrians' facing directions.

With the blossom of deep learning, CNN is well developed to extract the trajectory of a single object~\cite{hong2015online, ma2015hierarchical,wang2016stct}. Chu et al. developed STAM to detect more objects~\cite{chu2017online}. Fang et al. improved the performance of tracking by using RNN~\cite{fang2018recurrent}. The authors in\cite{wojke2017simple} developed the SORT model to track pedestrians. However, by tightly coupling detection and tracking, these approaches cannot always provide satisfactory performance in pedestrian detection. Mask R-CNN~\cite{he2017mask} and YOLO~\cite{redmon2016you} are two state-of-the-art detection networks with sufficient performance for detecting purposes, where YOLO is much faster than Mask R-CNN, and thus is more suitable for real-time tracking tasks.

\subsection{Navigation in Urban Environments}
Compared to the fixed video surveillance system, the surveillance robot not only has the above perception capabilities but also endowed the surveillance camera with mobility. However, navigating a robot in urban environments is non-trivial. 

First, the robot would inevitably interact with dynamic obstacles like pedestrians, bicycles. Some studies have been proposed to deal with the collision avoidance problems in such dynamic scenarios. \cite{van2008reciprocal,van2011reciprocal} proposed that each agent in dynamics scenarios should take half of the responsibility of collision avoidance. Based on that, they develop the multi-agent collision avoidance algorithm with zero-communication. \cite{Trautman:2010:IROS,Trautman:2013:ICRA} presented the interacting Gaussian processes to capture the cooperative collision avoidance behavior, and introduced the cooperative planner for robot navigation. However, these algorithms fail to track a moving pedestrian without the assistance of external devices. \cite{chen2017socially,everett2018motion} deployed a LiDAR with multiple cameras on robot to track surrounding pedestrians. To navigate the robot in the crowds, they utilized the reinforcement learning algorithm to train the socially aware collision avoidance policy. Different from the above algorithms, \cite{long2018towards,fan2018ijrr,fan2019getting} proposed a sensor-level collision avoidance policy learned via reinforcement learning, which can directly process the raw LiDAR data to generate collision-free actions.

\begin{figure}
    \centering
    \begin{subfigure}{0.47\textwidth}
    \includegraphics[width=1.0\linewidth]{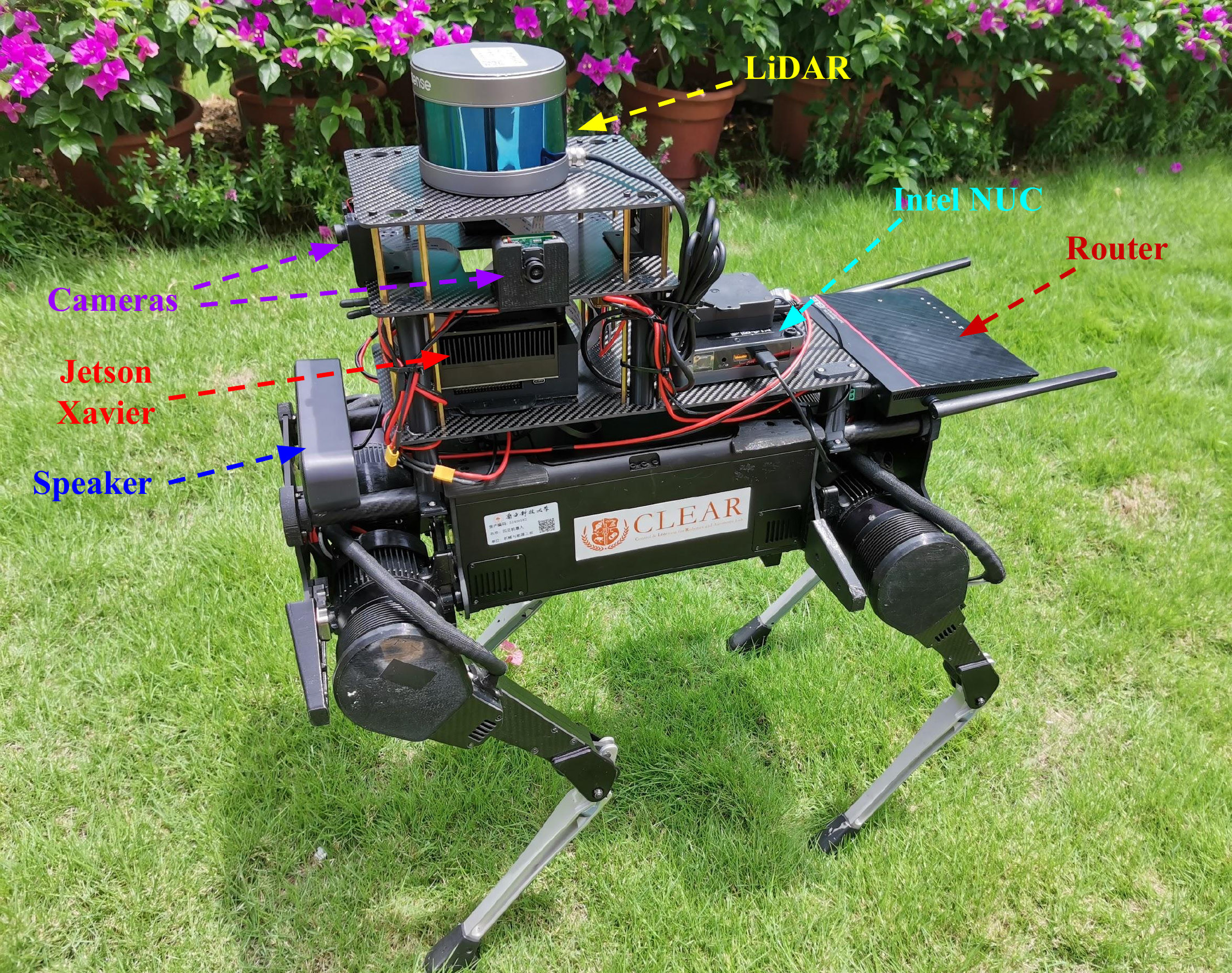}
    \label{fig:overview_dog}
    \end{subfigure}
    \begin{subfigure}{0.47\textwidth}
    \includegraphics[width=1.0\linewidth]{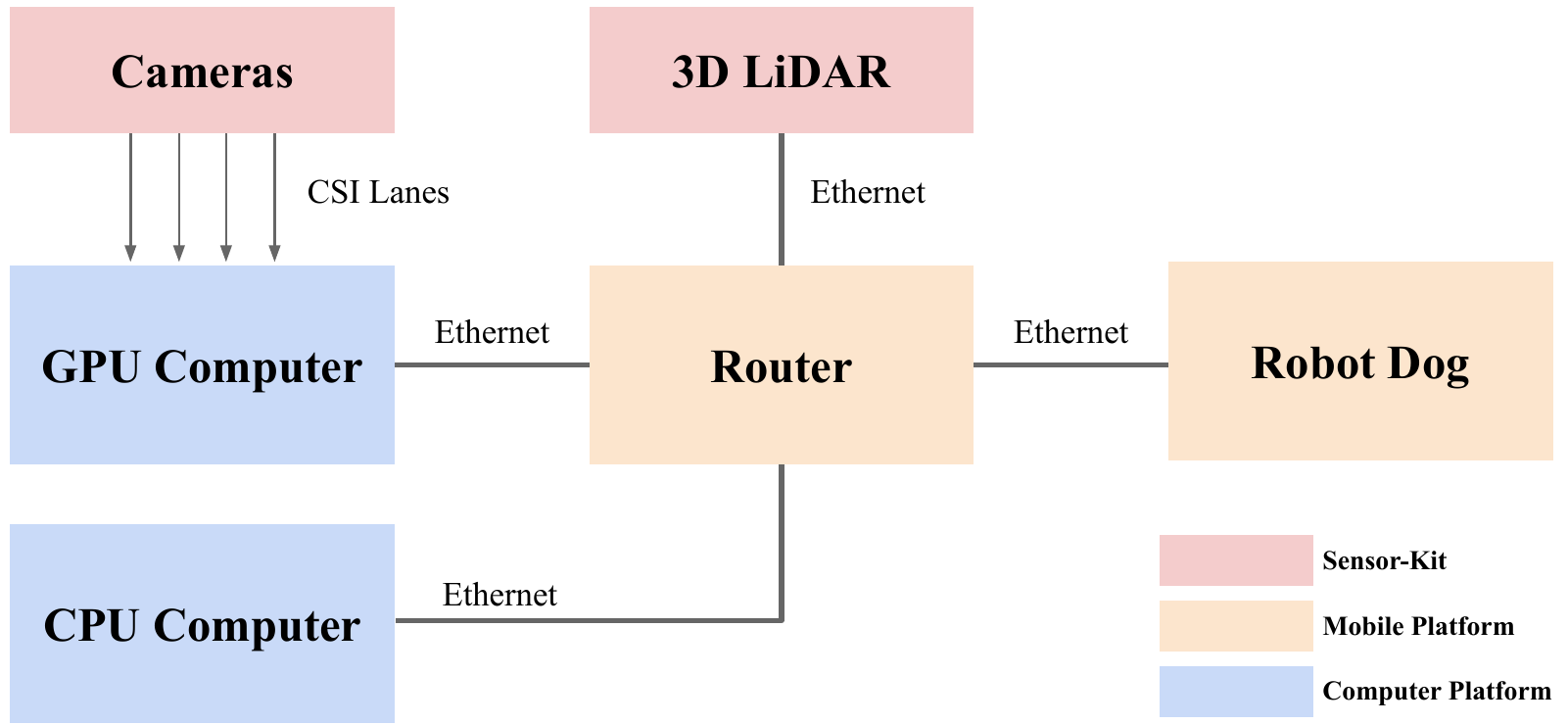}
    \label{fig:hardware_link}
    \end{subfigure}
    \caption{Overview of our hardware and software system.}
    \label{fig:hardware}
\end{figure}

\begin{figure*}[t]
    \centering
    \includegraphics[width =\linewidth]{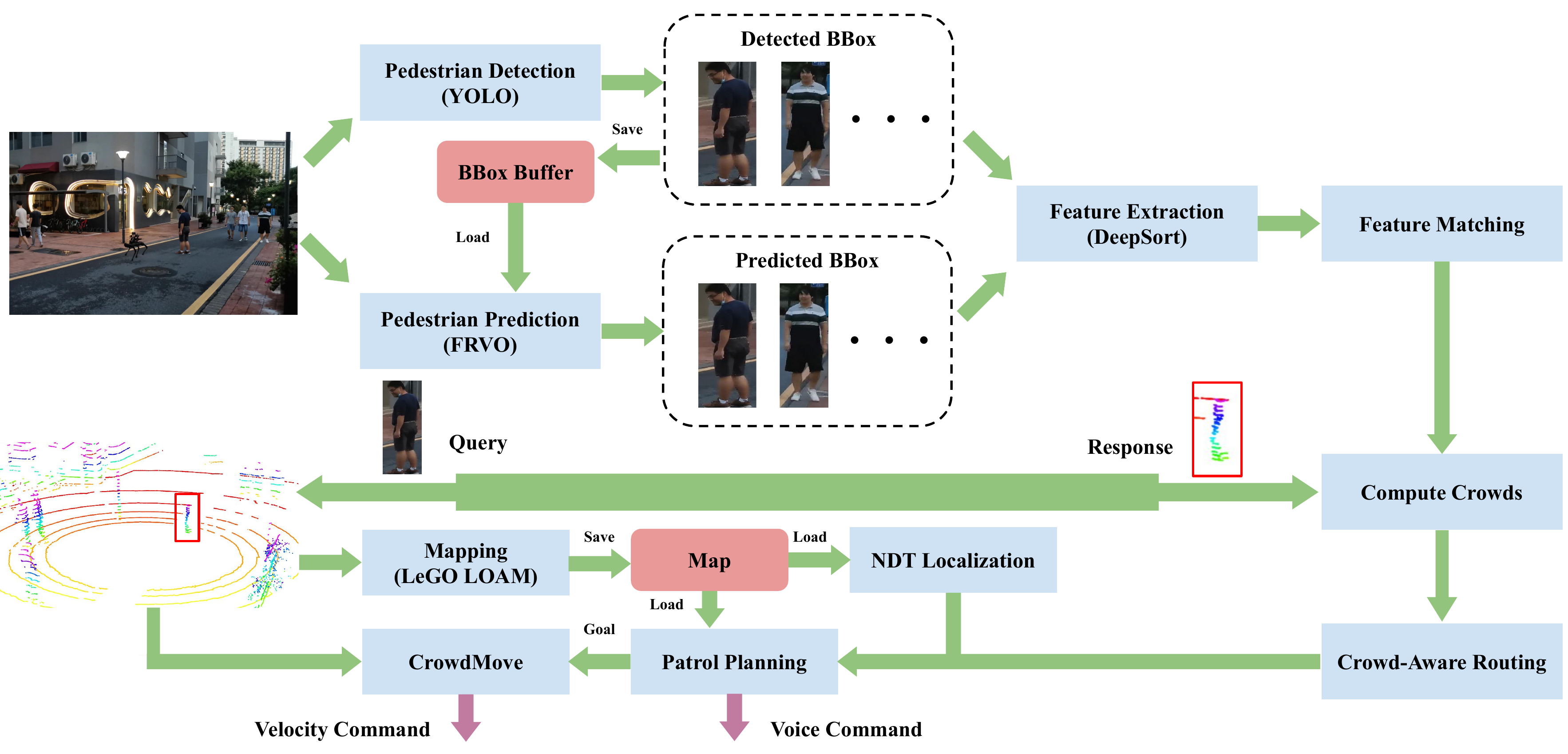}
\caption{
Our system contains functional modules of tracking, mapping, localization, patrol planning, routing, and motion planning. Tracking module uses YOLO~\cite{redmon2016you} and F-RVO~\cite{chandra2019densepeds} to extract similar detected objects of consecutive frames and to keep track of people. Mapping is achieved by using LeGO-LOAM algorithm which is based on 3D Lidar sensor. For localization, we used NDT localization algorithm to match lidar data and localize robot in the generated map. According to the detected crowds and map, crowd-aware routing algorithm and patrol planning algorithm would help robot to determine current target to approach. With all information needed for motion planning, an end-to-end algorithm, CrowdMove, is used to drive robot toward the goal position. During the approaching, if the robot detects its distance to the crowds is lower than 5 meters, it starts to play a recorded vocal command to remind people to keep a proper social distance.}
\label{fig:data_stream}
\end{figure*}

\subsection{Robot Interaction}
Human-like characteristics of social robots would influence users' response. Among various social traits, gender is important for interpersonal relationships and evokes social stereotypes~\cite{muscanell2012make}. Previous research has pointed out that the participants were more accepting of the robots if their perceived gender of a robot conformed to their occupation's gender role stereotypes (e.g., male security robots or female healthcare robots). However, perceived trust of the social robots was not influenced by gender-occupational role conformity~\cite{tay2014stereotypes}. In contrast, Kuchenbrandt \textit{et al.}~\cite{kuchenbrandt2014keep} found that participants, regardless of gender, evaluated the male and female robots as equally competent while performing a stereotypically female task but, in the context of a stereotypically male task, the female robot was rated as more competent compared to the male robot. Another study examining the effects of robot gender on human behavior found that participants were more likely to rate the robot of the opposite gender as more credible, trustworthy, and engaging~\cite{siegel2009persuasive}. Thus, the effects of users and robot attributes, as well as gender-role stereotypes, are still open questions. 

\section{Hardware Platform}
\label{sec:hardware}

First, we will introduce the hardware setup of our surveillance robot, which includes three components as shown in \prettyref{fig:hardware}: the mobile platform, the perception sensor-kit, and the computational platform.

\begin{itemize}
    \item \textbf{Mobile Platform}: We deploy the Laikago (a dog-like legged robot) as our mobile platform for navigating in complex urban environments. Comparing to the wheeled robot, the legged robot has superiority on traversability and thus is more suitable for uneven and unstructured urban scenarios with stairs and bumps. 
    \item \textbf{Perception Sensor-Kit}: To effectively detect and track pedestrians, we mount four color cameras evenly in the horizontal plane of the robot. Each camera is equipped with a short focal lens with the horizontal FOV of \SI{80}{^o}, and thus a combination of four cameras can almost cover all directions around the robot. Moreover, for better spatial perception, we use a robosense 3D LiDAR with 16 channels to measure the social distance between pedestrians. The 3D LiDAR also serves the navigation applications in mapping, localization, and collision avoidance.  
    \item \textbf{Computational Platform}: Two on-board computers are mounted to process the aforementioned sensor data for different tasks. We use NVIDIA Jetson AGX Xavier as the vision computational module that supports a maximum of six lanes CSI cameras as the input and uses 512 CUDA cores to GPU-accelerate the processing of images captured by the cameras. Since other tasks like mapping and localization would mostly consume CPU resources, we also deploy an Intel NUC with Intel i5 8259U CPU. These two computers are connected by wired network, and the processed data is shared by Robotic Operating System (ROS). 
    
\end{itemize}


\section{Social Distancing Detection}
\label{sec:tracking}

\begin{figure*}[h]
    \begin{subfigure}{0.52\textwidth}
    \includegraphics[width=1.\linewidth]{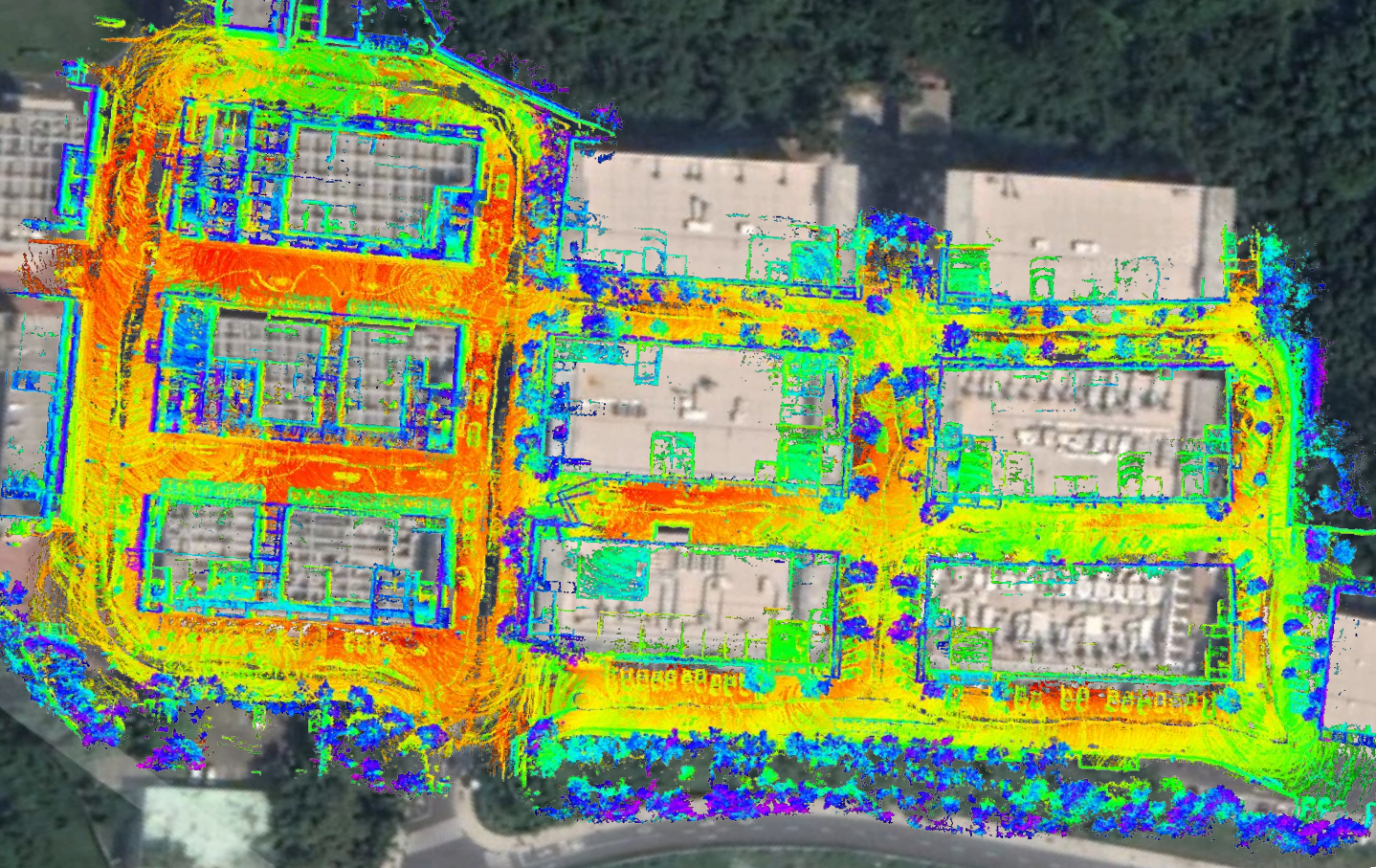}
    \caption{Mapping: LeGO-LOAM}
    \label{fig:mapping}
    \end{subfigure}
    \begin{subfigure}{0.48\textwidth}
    \includegraphics[width=1.\linewidth]{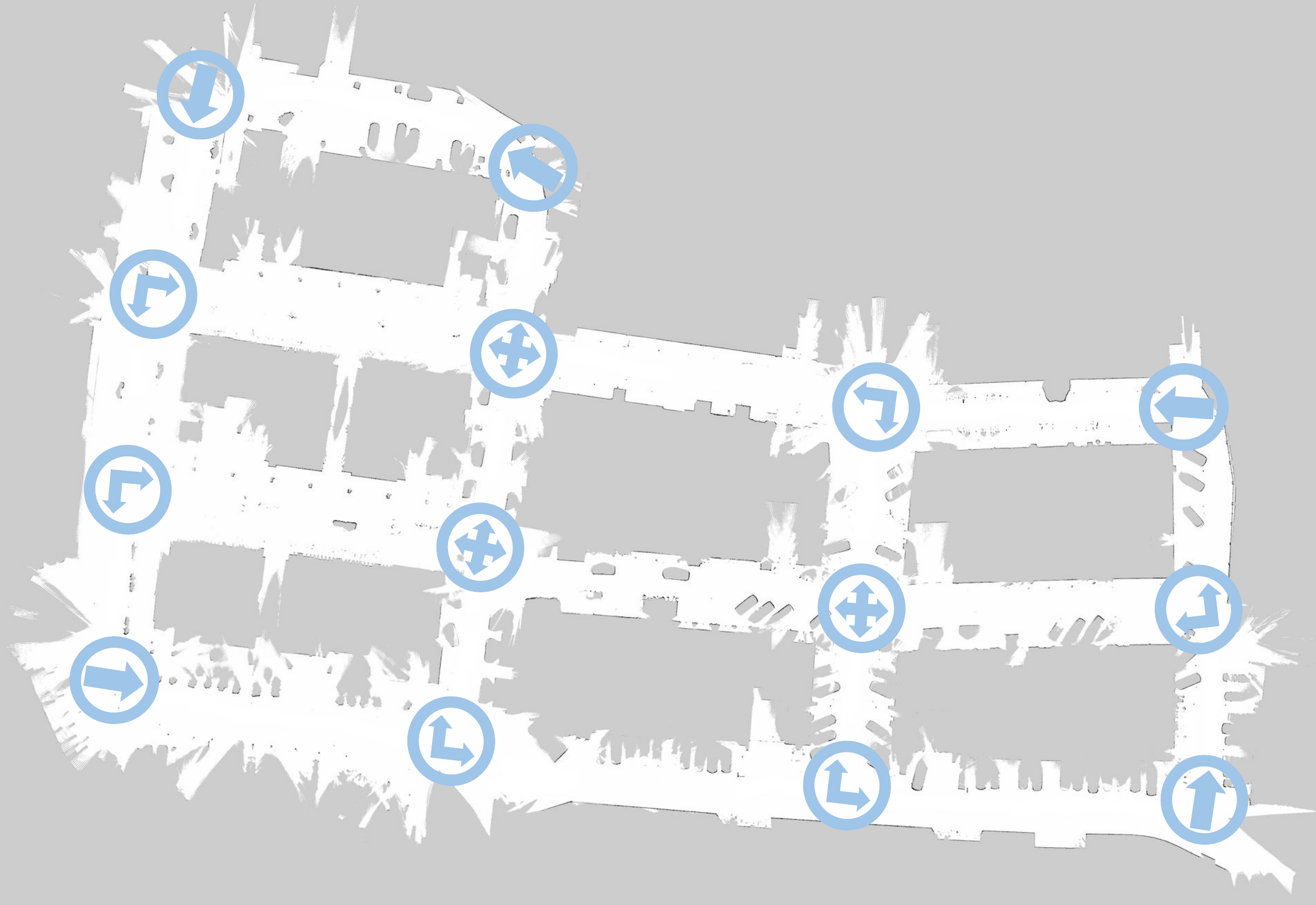}
    \caption{Potential directions in each cross and corner}
    \label{fig:direction}
    \end{subfigure}
    \caption{(a) We used LeGO-LOAM for mapping. The blue arrows in (b) shows the potential directions in each cross and corner that robot can go. Based on the information of the map and current position, the patrol algorithm chooses for the robot a navigation direction with the maximum probability of crowd appearance.}

    \label{fig:mappinwg_patrol}
\end{figure*}

The tracking algorithm used in our system composes of object detection, bounding box prediction, feature extraction, and sparse feature matching. We use YOLO to detect pedestrians, and match sparse features with help of motion modeling algorithm F-RVO to update the traces of pedestrians.


\subsection{F-RVO}
Modeling pedestrians behavior in crowds from the front view is challenging, not only because of the non-linear varied motions (turning shoulder, side walking, back stepping, etc.~\cite{best2016real}), but also due to the occlusions that front view may encounter. In this work, we use a velocity-obstacle based algorithm, F-RVO~\cite{chandra2019densepeds}, to model the pedestrians motion.

In F-RVO, each pedestrian, $p_i$, is represented by an 8-dimensinal vector: $\Psi_t = \left[ x, v, v_{pref}, l, w \right]$, where $x$ is the current position, $v$ is the velocity, $v_{pref}$ is the preferred velocity that we assume people would prefer to walk along the front direction. $l$ and $w$ are the length and width of human's shoulder. For each frame $\tau$, a half-plane constraint is used to determine the range parameter in F-RVO. Within the range, each pedestrian $p_i$ has an area of velocity obstacle $VO^{\tau}_{p_i | p_j}$ with respect to another neighbor pedestrian $p_j$. The convex region of velocity obstacles considering all neighbors can then be computed as:
\begin{equation}
FRVO^{\tau}_{p_i} = \bigcup_{p_j \in H_i} VO^{\tau}_{p_i | p_j},
\label{eqn:vo}
\end{equation}
where $H_i$ is the set of all neighbors of pedestrian $p_i$. Out of the velocity obstacle area, the best velocity is chosen with the nearest distance to preferred velocity $v_{pref}$:
\begin{equation}
v_{best} = \arg\min_{v} {\left\|  v - v_{pref}\right\| },
\label{eqn:velocity}
\end{equation}
where $v \notin  FRVO^{\tau}_{p_i}$.

\subsection{DensePeds}
The tracking algorithm, DensePeds, includes three components to track pedestrians: object detection, feature extraction, and feature matching, as shown in upside of \prettyref{fig:data_stream}. In each time step, We firstly use YOLO to detect pedestrians and generate bounding boxes for them. These detected pedestrians make a set $P$. Then we use F-RVO to predict another set of bounding boxes around the pedestrians $p_i \in P$. Given the bounding boxes computed in two adjacent time steps, we use DeepSort CNN~\cite{wojke2017simple} to extract binary feature vectors from the sub-images as determined by the bounding boxes. Then we perform matching over these sparse features to find in frame $t+1$ the best matched pedestrians of frame $t$ and assigned IDs to the pedestrians accordingly. In particular, the sparse features are matched in two steps. First, we find the most similar detected pedestrian of a predicted pedestrian using the cosine metric, i.e., 
\begin{equation}
h^{*}_{j} = \arg \min_{h_j}\{d(f_{p_i}, f_{h_j}) \mid p_i\in P, h_j \in H_i\}
\label{eqn:cosin}
\end{equation}
where $d(\cdot,\cdot)$ is the cosine metric, $f(\cdot)$ is the feature extraction function, $p_i$ is one pedestrian in the set $P$ of all pedestrians in a frame, $h_j$ is one detected pedestrian in the set $H_i$, which is the set of detected neighbors around the pedestrian $p_i$. In the second step, we maximize the IoU overlap, i.e., the overlapped area between predicted boxes and original YOLO-detected boxes, 
\begin{equation}
\epsilon(i,j) = \frac{B_{p_i} \cap B_{h_j}}{B_{p_i} \cup B_{h_j}},
\label{eqn:iou}
\end{equation}
where $B_{p_i}$ and $B_{h_j}$ are the bounding boxes around $p_i$ and $h_j$ respectively. Matching a set of detected pedestrians to a set of predicted pedestrians with maximum overlap eventually becomes a max weight matching problem over the matrix $\epsilon(i, j)$, which can be accelerated using the Hungarian algorithm~\cite{kuhn1955hungarian}.

According to the computed bounding boxes, we can roughly estimate the range and bearing information between the robot and pedestrians. To more accurately estimate the crowds, we reproject the bounding boxes to the LiDAR coordinate to query the depth of each pedestrian. The RANSAC algorithm\cite{RANSAC_ALGORITHM} is applied to filter out the possible outlier points. If there is a large inconsistency between LiDAR and visual estimates due to the occupation between pedestrians, the visual estimates would be trusted. Finally, we obtain the social distance between pedestrians.

\section{Navigation in Urban}
\label{sec:nav}

In this section, we will introduce the autonomous navigation algorithm for urban scenarios. We implement the mapping and localization function by the state-of-the-art LiDAR-based approaches. The navigation framework is formulated as the hierarchical structure. In particular, we develop a learning-based collision avoidance algorithm for local planning, and use global planning to plan trajectories for robots to patrol. In addition, we will describe the routing algorithm enabling a robot to effectively select a crowded region to approach in order to accomplish the patrol. 

\subsection{Mapping and Localization}
Since LiDAR-based SLAM approaches have been well developed in recent years, we are not going to develop a new SLAM approach for this paper. To achieve the best performance for this legged robot, we choose the LeGO-LOAM algorithm, which is a light-weighted system and is optimized for the grounded platform~\cite{LeGO-LOAM}. The generated map is shown in \prettyref{fig:mapping}. 

After the robot obtained the 3D point cloud map about the scenario, the Normal Distributions Transform (NDT) scan matching algorithm is used for localization~\cite{NDT-MATCHING}, which have been demonstrated in~\cite{3D_MATCHING_COMPARE} to be able to provide more reliable result than other matching methods such as Iterative Closest Points~\cite{ICP}. 

Although we can compute the 3D point cloud map and the robot's localization, it is not easy for the robot to determine the traversable region in the 2D plane. Therefore, we transform the 3D point cloud to the 2D laser scan, by taking the closest point within the certain height as a 2D laser point. Note that, during the navigation the robot may encounter uneven terrains like stairs or steps. Thus, the transform ignores the point cloud on the ground plane by filtering out the cloud points lower than \SI{30}{cm}. 
After the transform, we obtain a 2D occupancy map for the following navigation algorithm as shown in \prettyref{fig:direction}.

\begin{figure}
\centering
\includegraphics[width=1.\linewidth]{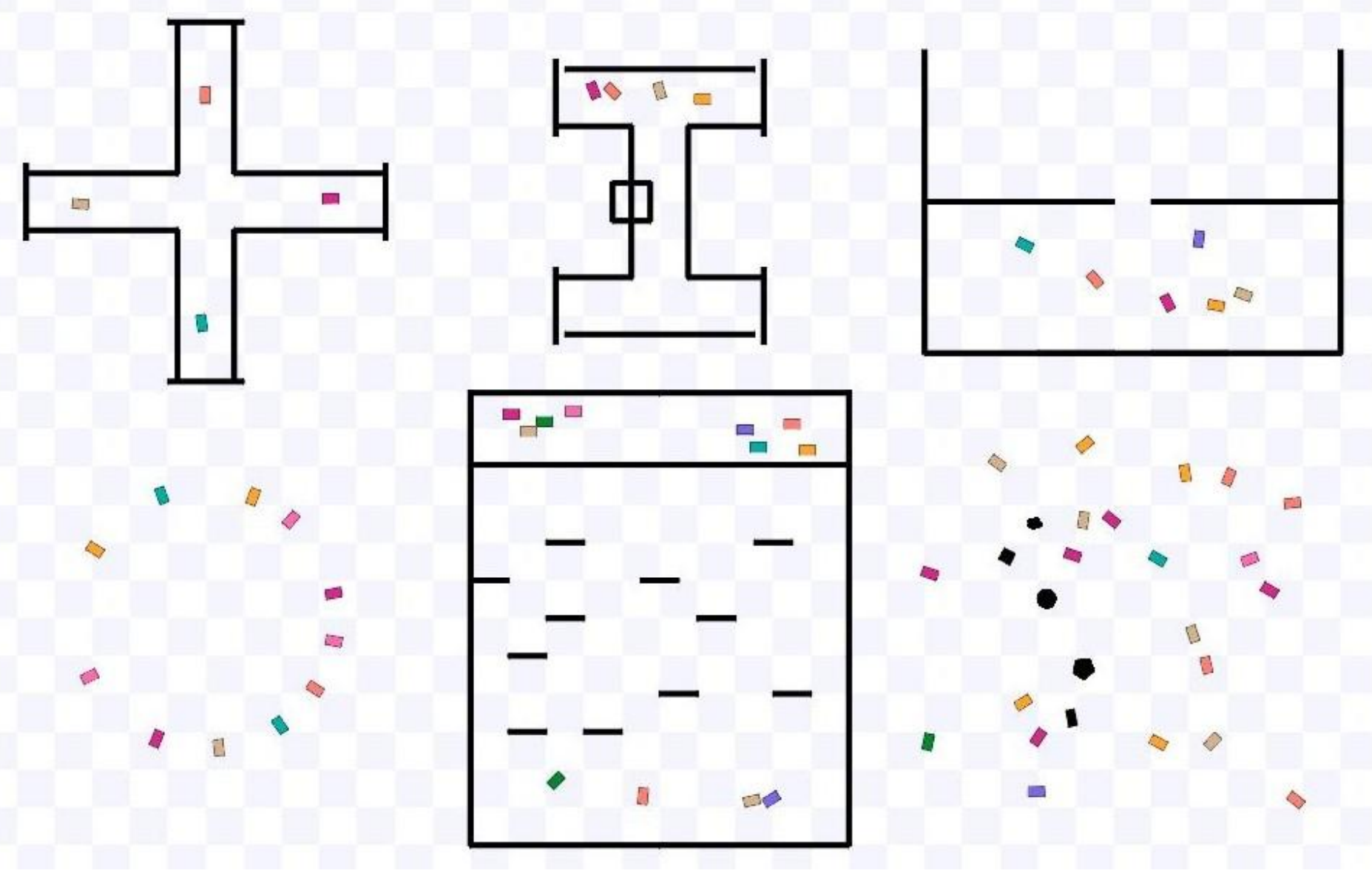}
\caption{Multi-robot multi-scenario training environments in the Stage simulator}
\label{fig:stage}
\end{figure}

\subsection{Patrol and Routing}
Based on the generated map and current position, we proposed a patrol planning algorithm to navigate robot traveling around the mapped area. As shown in \prettyref{fig:mappinwg_patrol}, in different crosses and corners, the robot would choose different navigation directions optimal for social distancing. In particular, the robot would prefer the direction where there is a high probability that a crowd would appear. 

When the robot detected gathered crowds, it would suspend the patrol algorithm and switch to the routing algorithm to find an optimal way to approach the crowds. Considering the time constraints and size of crowds, we propose a crowd-aware routing algorithm based on the depth-first search method to find a sequence of intermediate waypoints for the robot to follow.

We formulate the routing problem as follows. Assume that there are $N$ groups of people within the robot's perception range. Each crowd is denoted as a node $n_i$, with its specific time-window constraint $t_i$, and its relative location to the robot. Each crowd is assigned a weight $w_i$ according to the number of persons in the group. The routing algorithm aims at finding an optimal path for the robot to approach as many crowds as possible with the least energy consumption. The optimization objective is:
\begin{equation}
cost = \sum_{p_j \in p_c}{e_j} - \sum_{i\in N}{w_i} - n_c,
\label{eqn:costfunction}
\end{equation}
where $p_c$ is the current trajectory which contains a set of points and edges denoting the positions of crowds and the paths connecting them. Each edge $p_j \in p_c$ between two positions has the energy cost $e_j$. The number of crowds explored in $p_c$ is denoted as $n_c$.

Given the directions and positions of the crowds after routing algorithm, we implement the SBPL lattice planner~\cite{SBPL_LATTICE_PLANNER} to generate a smooth patrol route passing through these way-points.

\subsection{Learning-based Collision Avoidance}
During patrol, the robot will not only encounter the static obstacles, but also interact with moving pedestrians. For this case, we deployed the learning-based collision avoidance approach, CrowdMove\cite{fan2018crowdmove}, for robotic navigation in crowds. 

The main training framework refers to our previous work~\cite{long2018towards}, which takes a 2D laser scan as the input and outputs the velocity command. The multiple training scenarios are designed with multiple robots in the Stage simulator as shown in~\prettyref{fig:stage}. We introduce the \textit{centralized learning, decentralized execution} training paradigm, which shares the same navigation policy during the training. Then, we obtain a multi-robot collision avoidance policy with zero communication. Furthermore, we validate that the trained policy can be transferred from the simulation to the real world without any re-tuning, and it is also suitable for the single robot navigation in crowds~\cite{fan2018ijrr,fan2019getting}. To make the training framework work for our hardware platform, we take the transformed laser scan which represents the local traversable area as the input.

\section{Voice Interaction}
\label{sec:interaction}

In our surveillance scenario, we use verbal cues to send suggestions from robot to human. As we mentioned before, the user's gender and the robot's gender may influence the user's acceptance and trust in the robot. Thus, to reach an effective surveillance result, we gave our robot four types of gendered voice and designed a user study to select the best one. In this section, we introduce the study for investigating (1) the user's gender-based effects of the autonomous robot and, (2) user's attitude, acceptance, trust, and perceived trust through robot with different voices.

\subsection{Method}
\subsubsection{Gender of the robot}
We manipulated the gender of the robot through non-verbal cues by changing the vocal characteristics. Because we aim to find the robot voice with best performance, the voice selection is not strictly limited to robot gender effects. In this experiment, we prepared four types of voices: three gendered voices and a child voice. The gendered voices include a computer-generated neutral voice, a male and a female recorded by real adult human, a child voice by a girl.

\begin{table*}[hbt]
\caption{Dependent measures in the user questionnaire}
\label{tab:matrics}
\resizebox{\textwidth}{!}{
\begin{tabular}{lcccl}
\hline
Factors                & \# Items & Cronbach Alpha & Scale        & Items                                                                                                                                                                                                                                                                                                                                      \\\hline
Attitude toward robots & 2            &  0.836     & Likert scale & \begin{tabular}[c]{@{}l@{}} It's a good idea to use the surveillance robot.\\ It's good to make use of the robot\end{tabular}                                                                                                                                                                          \\\hline
Perceived trust        & 3            &  0.835     & Likert scale & \begin{tabular}[c]{@{}l@{}}I would trust the surveillance robot, if he/she gave me advice.\\ I trust that the surveillance robot can keep me away from health risks.\\ I would follow the advice that the surveillance robot gives me.\end{tabular}                                                          \\\hline
Acceptance             & 3            &  0.868     & Likert scale & \begin{tabular}[c]{@{}l@{}}If given a chance, I'll use this robot in a college campus in the near future. \\ If given a chance, I'll use this robot in a park in the near future.\\ If given a chance, I'll use this robot in a shopping mall in the near future.\end{tabular} \\\hline
Perceived ability      & 3            &  0.864      & Likert scale & \begin{tabular}[c]{@{}l@{}}The robot is very capable of performing its job. \\ I feel very confident about the robot's skills. \\ The robot has much knowledge about the work that needs to do.\end{tabular}                                                                                               \\     \hline

\end{tabular}
}
\end{table*}

\subsubsection{Procedures}
Among the various issues in human-robot interaction, trust was nominated as one of the primary factors to be considered. In this particular task, trust is performed as how much a human would follow the advice sent by the surveillance robot. This factor would crucially influence the performance of the robot. To better measure the users' experience of the robot, we suggest four dependent measures which include the users' attitude towards the robot, perceived trust, and acceptance of the robot. The details of the measures are shown in Table~\ref{tab:matrics}.


As part of a larger study investigating the users' perceptions in an autonomous surveillance robot, the participants filled out a survey measuring the factors shown in Tabel~\ref{tab:matrics}. Each measure was assessed on a 5-point Likert scale (‘1’ = strongly disagree, ‘5’ = strongly agree). 

\begin{figure}
    \centering
    \begin{subfigure}{\linewidth}
         \centering
         \includegraphics[trim=0 0 0 0, clip, width=\textwidth]{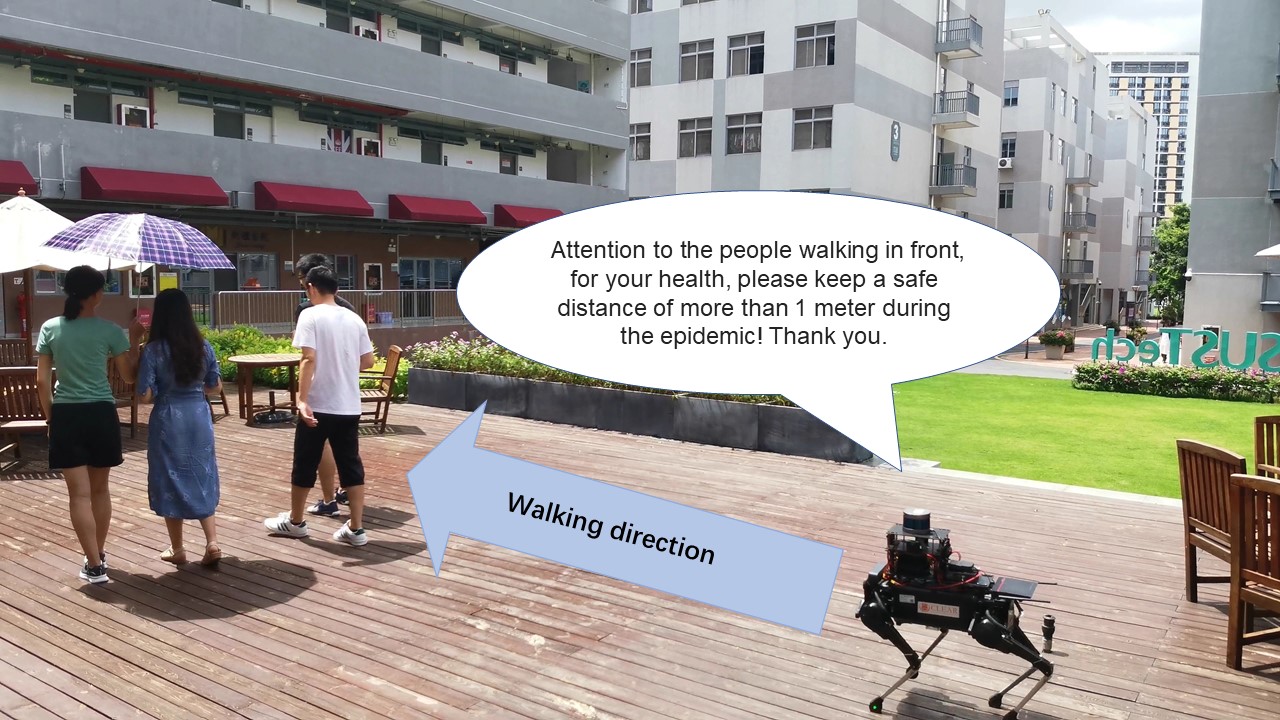}

    \end{subfigure}
    \vskip\baselineskip
    \begin{subfigure}{\linewidth}
           \centering
            \includegraphics[trim=0 0 0 0, clip,width=\textwidth]{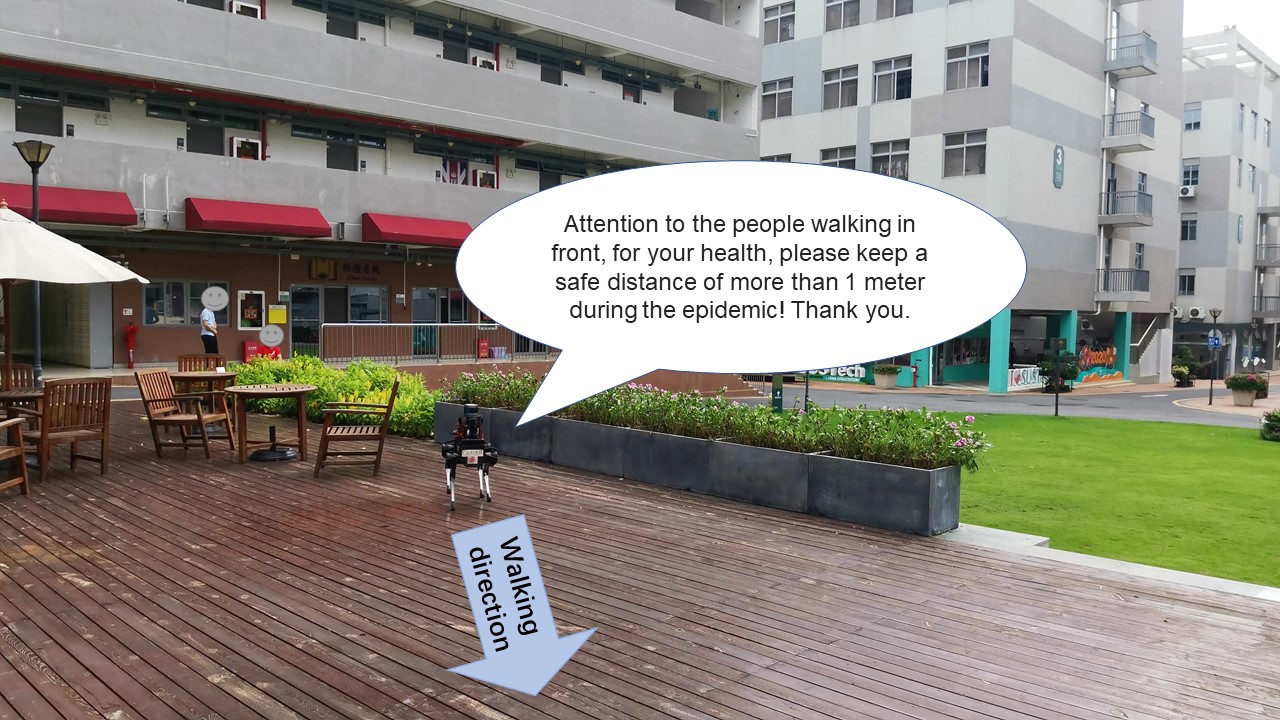}
           
    \end{subfigure}
    \caption{The screenshots of two videos in the questionnaire. Top: third-person perspective; bottom: first-person perspective.}
    \label{fig:video_view}
\end{figure}

This experiment was done in between-subject mode to minimize the learning and transfer across conditions. Each participant in this study viewed two videos and then responded to survey items related to the videos. Both of the videos demonstrate the same scenario with the same robot voice. One video was from a third-person perspective, where the robot is walking towards the crowds while asking them to keep the social distance. The other video was recorded from a first-person perspective where the robot is walking towards the camera while asking the human to keep the social distance. For both scenarios, the robot starts to play the voice at about 5 meters away from the crowd. The screenshots of the two videos are shown in figure~\ref{fig:video_view}. In this way, a total of 8 videos were recorded, which are 2 perspectives times 4 types of robot voices. For each scenario, we add a description ``The robot shown in the videos is a surveillance robot working on keeping a low density of humans during COVID-19. When the robot finds a crowd, he/she/it will walk toward the crowd while asking them to keep a proper social distance. Please watch these two videos, and imagine you were one of the humans in the video, then answer the following questions.''

\subsubsection{Participants}
A total of 218 adults (119 males; 99 females) between 20-55 years old (M=29.49, SD=12.02) participated in the between-subject experiment. Participants were mostly students and staff from the Southern University of Science and Technology. The participants were recruited through the posters and links shared in a social media app. Each participant needs to read and sign a consent form before they start the questionnaire. 

\begin{table*}[hbt]
\centering
\caption{Means and standard deviations of all the measures}
\label{tab:overall_result}
\resizebox{\textwidth}{!}{
\begin{tabular}{l|p{1cm}p{1cm}p{1cm}|p{1cm}p{1cm}p{1cm}|p{1cm}p{1cm}p{1cm}|p{1cm}p{1cm}p{1cm}}
                       & \multicolumn{3}{l}{Male voice}    & \multicolumn{3}{l}{Female voice}                         & \multicolumn{3}{l}{Neutral voice}                                                                       & \multicolumn{3}{l}{Child voice}                                                       \\\hline 
                       & overall (n=54)     & male (n=23)        & female (n=31)     & overall (n=53)    & male (n=31)                               & female (n=22)      & overall (n=56)                           & male (n=31)                                & female (n=25)                             & overall (n=55)                          & male (n=34)                              & female (n=21)     \\\hline 
Perceived ability      & 3.90 (0.78) & 3.62 (0.96) & 4.10 (0.56) & 4.05 (0.81) & 3.98 (0.82)                        & 4.15 (0.80) & 4.06 (0.77)                        & 3.91 (0.79)                         & \cellcolor{red!20} 4.24 (0.72) & \cellcolor{yellow!20} 4.15 (0.66) & \cellcolor{blue!20} 4.12 (0.69) & 4.19 (0.64)               \\
Acceptance             & 3.67 (0.98) & 3.59 (1.23) & 3.72 (0.76) & 4.06 (0.80) & \cellcolor{blue!20}  4.05 (0.83) & 4.08 (0.78) &\cellcolor{yellow!20}  4.08 (0.73) & 3.89 (0.74)                         &  \cellcolor{red!20}  4.32 (0.65) & 4.07 (0.72)                        & 4.02 (0.73)                        & 4.14 (0.70)   \\
Perceived trust        & 3.91 (0.60) & 3.86 (0.78) & 3.95 (0.43) & 4.09 (0.73) & 4.02 (0.72)                        & 4.18 (0.73) & \cellcolor{yellow!20}  4.20 (0.67) & 4.08 (0.69)                         & \cellcolor{red!20}  4.36 (0.62) & 4.13 (0.72)                        & \cellcolor{blue!20}4.10 (0.77) & 4.19 (0.66)  \\
Attitude toward robots & 3.96 (0.98) & 3.78 (1.10) & 4.10 (0.88) & 4.25 (0.79) & \cellcolor{blue!20}  4.21 (0.74) & 4.32 (0.87) & \cellcolor{yellow!20}  4.31 (0.66) &  4.16 (0.68) & \cellcolor{red!20}4.50 (0.61) & 4.15 (0.73)                        & 4.16 (0.71)                        & 4.14 (0.76)       \\\hline                    
\end{tabular}
}
\end{table*}


\begin{table*}[hbt]
\centering
\caption{F value and significance of robot voice effect and user gender effect}
\label{tab:significent}
\resizebox{0.8\linewidth}{!}{
\begin{tabular}{p{3.5cm}|p{20pt}p{20pt}|p{20pt}p{20pt}|p{20pt}p{20pt}|p{20pt}p{20pt}}

                       & \multicolumn{6}{l}{Robot voice effect} & \multicolumn{2}{l}{User gender effect} \\\hline
                       & overall    &           & male    &         & female   &           &                                &                                  \\
                       & F          & p         & F       & p       & F        & p         & F                              & p                                \\\hline
Perceived ability      & 1.027      & 0.382     & 1.761   & 0.158   & 0.221    & 0.882     & 5.160                          & 0.024**                          \\
Acceptance             & 3.362      & 0.020**   & 1.455   & 0.231   & 3.404    & 0.021**   & 1.285                          & 0.258                            \\
Perceived trust        & 1.866      & 0.136     & 0.563   & 0.640   & 2.214    & 0.092*   & 1.939                          & 0.165                            \\
Attitude toward robots & 2.017      & 0.113     & 1.534   & 0.209   & 1.397    & 0.248     & 2.071                          & 0.152   \\\hline    
 \multicolumn{9}{l}{\textit{* $p<0.1$, ** $p<0.05$}} \\\hline
\end{tabular}
}
\end{table*}

\subsection{Data analyses}
A manipulation check was performed to ensure that the robots could manifest gender and age successfully. The perceived gender was measured through a sliding bar with 0 the most femininity and 100 the most masculinity. The perceived age was measured through a sliding bar between 5 to 70. The one-way ANOVA showed that participants perceived male voice more masculine ($M=76.14$), female voice more feminine ($M=49.25$) an neutral voice in the middle ($M=64.28$). The F and p value is $F=9.902$ and $p<0.0001$. The participants also significantly perceived robot with child's voice ($M=20.93$) younger than others ($M=28.02$, $p=0.008$).

We calculated Cronbach's alpha values to assess the internal consistency of each psychometric measure. The reported alpha values were between 0.8-0.9, indicating that the items have relatively high internal consistency. To calculate the significance of user gender and robot voice type effect, a one-way ANOVA was conducted. The robot voice and user gender were treated as independent variables. For factors reached significant differences according to conditions, we used the least significant difference (LSD) to make a pairwise comparison.

\section{Experiments}
\label{sec:exp}

\begin{figure*}[t]
    \begin{subfigure}{0.62\textwidth}
    \includegraphics[width=1.\linewidth]{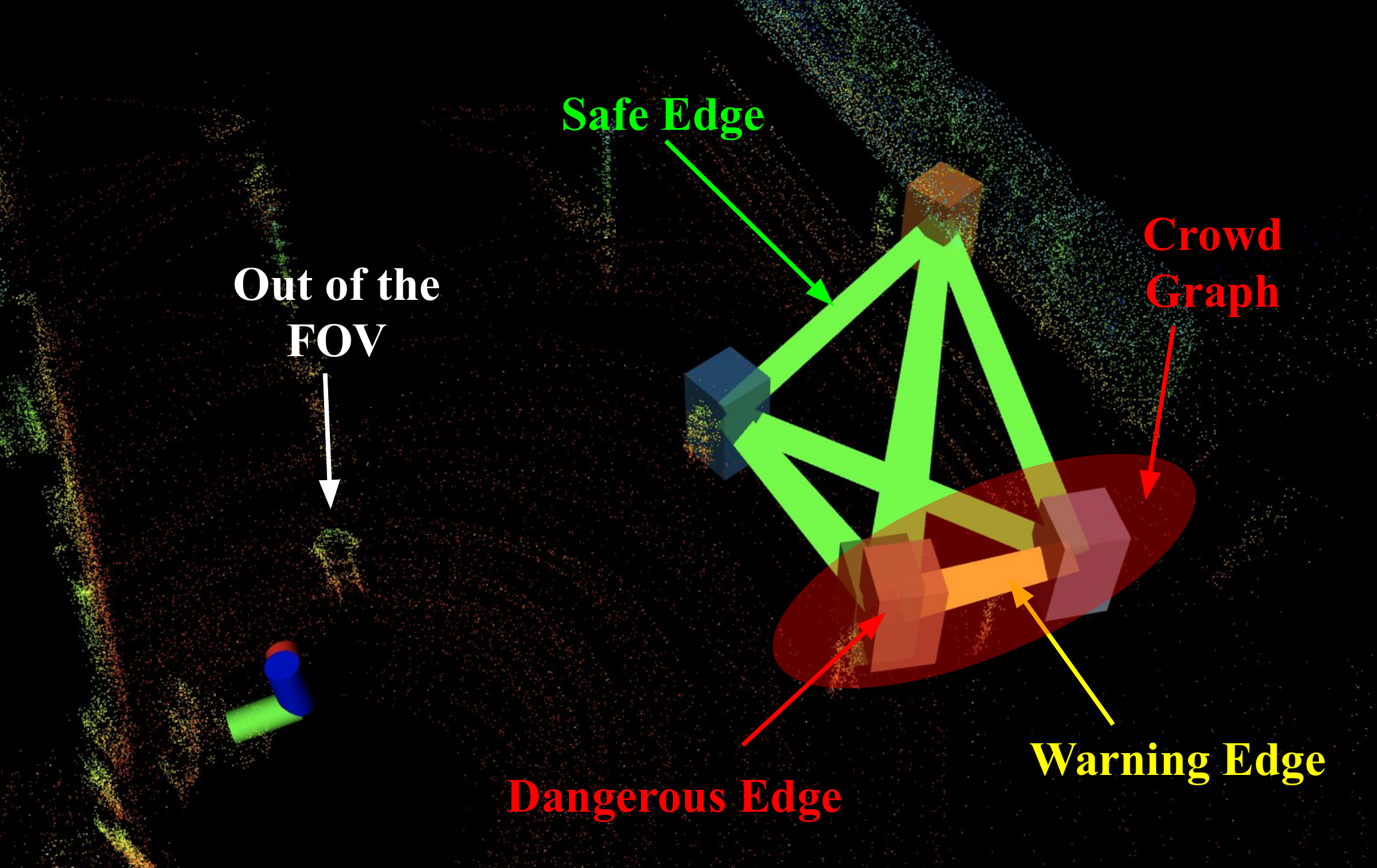}
    \caption{Visualization of the social distancing detection}
    \label{fig:ped_graph}
    \end{subfigure}
    \begin{subfigure}{0.37\textwidth}
    \includegraphics[width=1.\linewidth]{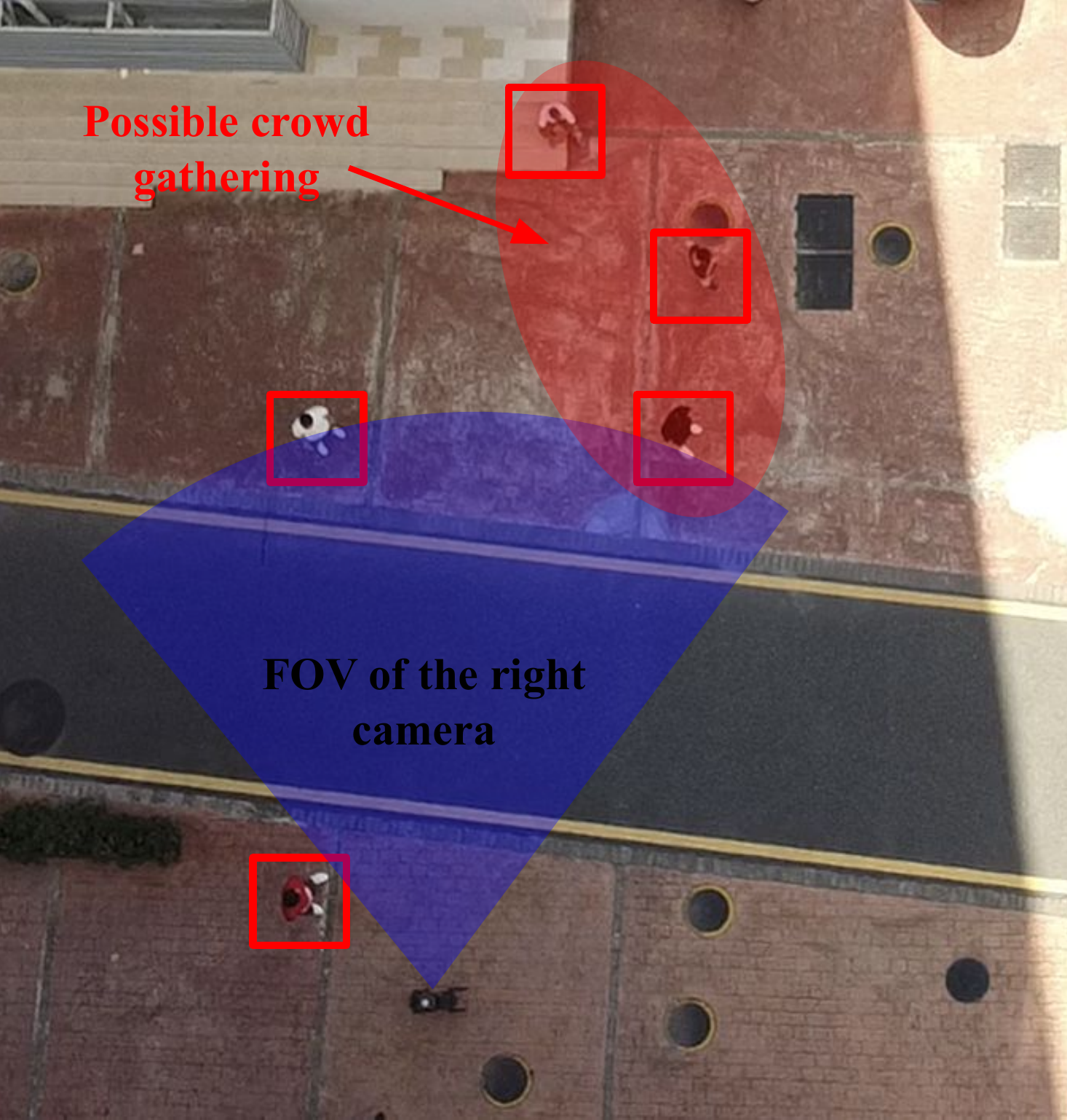}
    \caption{Bird view of the scenario}
    \label{fig:patrol}
    \end{subfigure}
    \caption{Illustration of the crowd gathering detection within the right camera's field of view (FOV). Although the estimated position of pedestrians is not very accurate, we can still detect possible crowd gatherings by establishing the crowd subgraph. }

    \label{fig:tracking_demo}
\end{figure*}

In this section, we first validate the effectiveness of the proposed approach individually. Then, we integrate all the modules to realize the autonomous surveillance robot. To further investigate the performance of surveillance robot on promoting social distancing, we conduct some real-world experiments in the end.  

\subsection{Crowd Gathering Detection}

We first record vision and LiDAR data to better analyze and tune the social distancing detection system. The recorded dataset includes a wide variety of pedestrian group behaviors, such as walking, standing, gathering, and scattering. 

Crowd gathering is not easy to be well quantified, especially the occlusion between pedestrians makes the robot difficult or even impossible to accurately acquire the location of each pedestrian. To detect each possible crowd gathering, we establish a graph-based pedestrian network called \textit{social graph}, with one example illustrated in \prettyref{fig:ped_graph}. In the social graph, each node represents the pedestrian's position. The green, yellow and red edges represent the safe, warning, and dangerous social distance, respectively. We connect the nodes between red and yellow edges into a subgraph 
called the \textit{crowd graph}, which is considered as possible crowd gathering. In this way, 
we can reduce the dependence of the crowd gathering detection on the accuracy of estimating pedestrian positions.

\subsection{Navigation in Urban}

\begin{figure}
\centering
\includegraphics[width=1\linewidth]{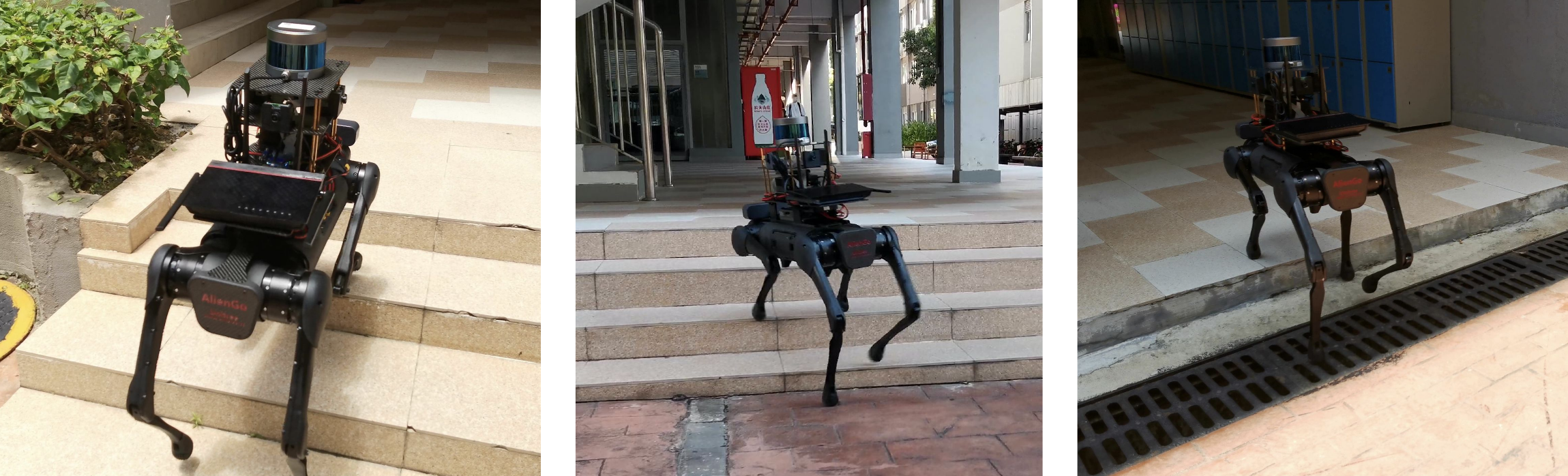}
\caption{The legged robot can traverse in uneven terrains.}
\label{fig:dog_steps}
\end{figure}

The urban navigation would mainly encounter two challenges, the unstructured environments and the dynamic obstacles. Thanks to the superior mobility of the quadruped platform, our robot can navigate over uneven terrains such as steps without extra visual estimation effort as shown in \prettyref{fig:dog_steps} and thus can handle unstructured environments easily.

\begin{figure}
\centering
\includegraphics[width=1\linewidth]{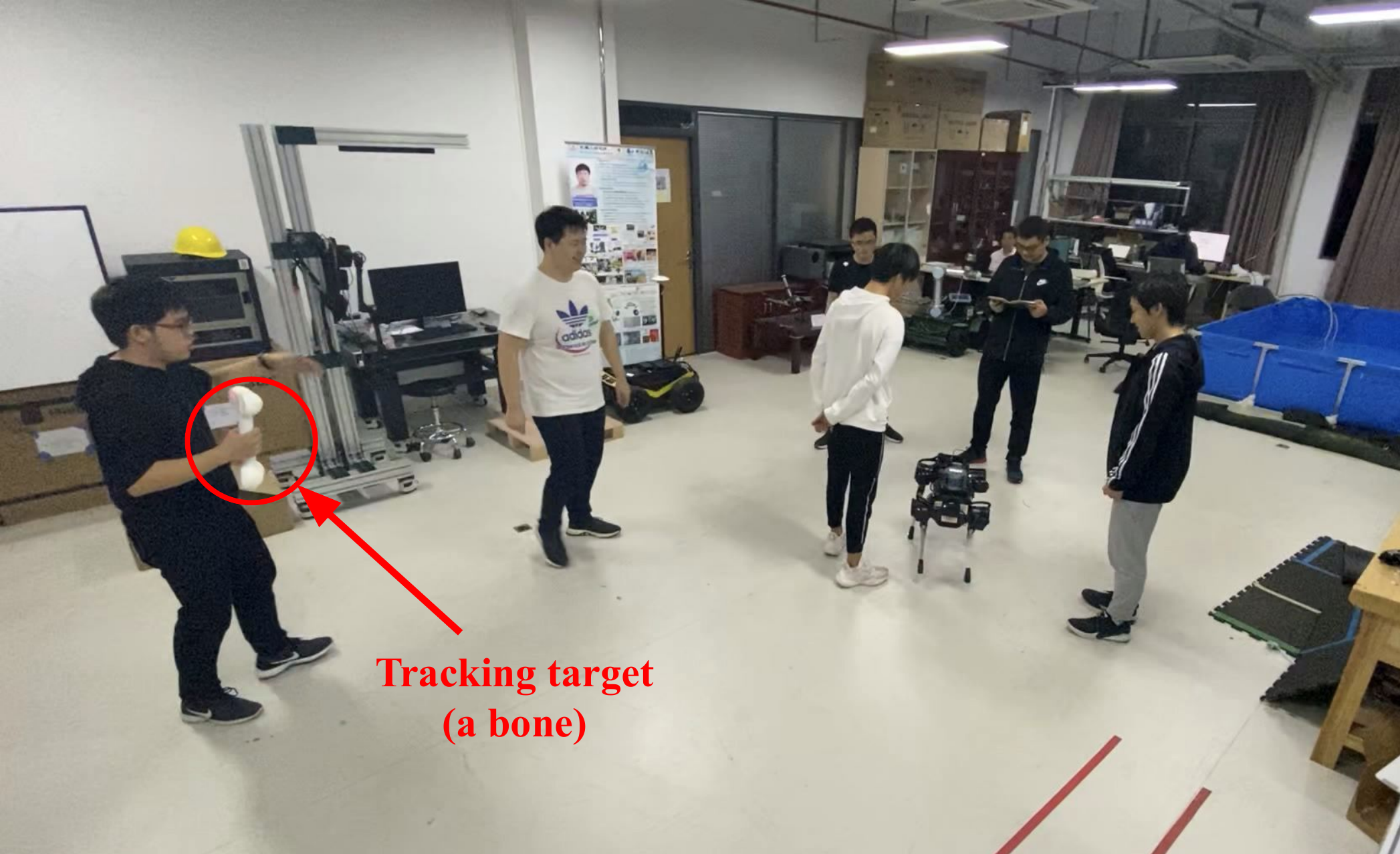}
\caption{The demonstration of the dynamic collision avoidance experiments. We arranged 6 moving pedestrians in this scenarios about \SI{4}{m} $\times$ \SI{4}{m} in size.}
\label{fig:dog_ca}
\end{figure}

To validate the dynamic collision avoidance performance among pedestrians, we create a crowded and narrow indoor scenario in the lab, as shown in \prettyref{fig:dog_ca}. In this experiment, the robot is required to perform tasks of tracking a specified target (a bone in this work). We install in the lab several ultra-wide band (UWB) tags accounting for indoor localization. During the experiments each lasting about 30 minutes, the robot dog mounted with a 3D LiDAR can achieve nearly zero collision in this scenario. This experiment indicates that our learning-based collision avoidance policy can be successfully transferred and deployed to the real-world robotic dog. 

\subsection{Voice Preference}
Table~\ref{tab:overall_result} shows the means and standard deviations of all measures according to different robot voice types and user genders. 
The score of each factor was calculated by averaging both/all the related items. It can be seen that the male voice type got the lowest average score on all the measures. Also, female users marked the highest on all the measures for neutral voice type.

Table~\ref{tab:significent} demonstrates the F and p-value. The result shows that for male users, there is no significant result among all different robot voice types while for female users, the robot voice type significantly influences the user's acceptance ($p=0.021$) and perceived trust ($p=0.092$). To find which condition differs for female users, We used the least significant difference (LSD) to make a pairwise comparison between different robot voice types. Surprisingly, For female users, the acceptance, perceived trust, and attitude toward robot in neutral robot voice condition are higher than other robot voice condition, especially for male voice condition ($p=0.003, 0.013, 0.061$ respectively).  

We also compare the effects of users' genders in different conditions. It is found that female user has higher perceived ability than male users ($p=0.024$), especially in male voice condition ($p=0.027$). In the neutral voice condition, female's acceptance and attitude toward robots are significantly higher than male's ($p=0.028$ and $p=0.057$ respectively).

There is no significant difference for male users markings according to different robot voice types. However, it is quite surprising that the female users marked very high for the neutral robot voice. In addition, surveillance should be a masculine job but both male and female users marked all four factors the lowest in male voice condition. Therefore, we do not suggest the usage of the male robot voice. Considering the means of the four factors among all robot voice conditions, we selected the neutral voice for our surveillance robot.

\subsection{Real-world experiment on Promoting Social Distancing}
\begin{figure*}[htb]
    \centering
    \begin{subfigure}{0.49\linewidth}
    \includegraphics[trim={10 10 10 10},clip,width=\textwidth]{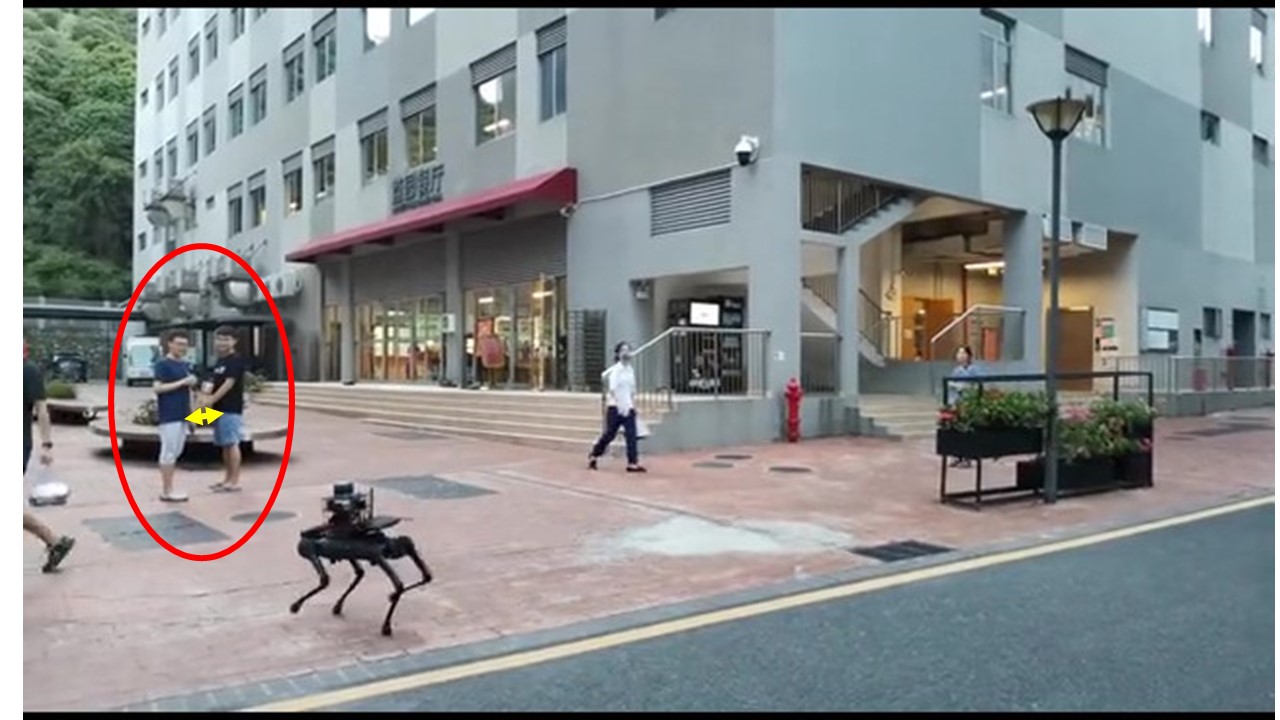}
    \end{subfigure}
    \begin{subfigure}{0.49\linewidth}
    \includegraphics[trim={10 10 10 10},clip,width=\textwidth]{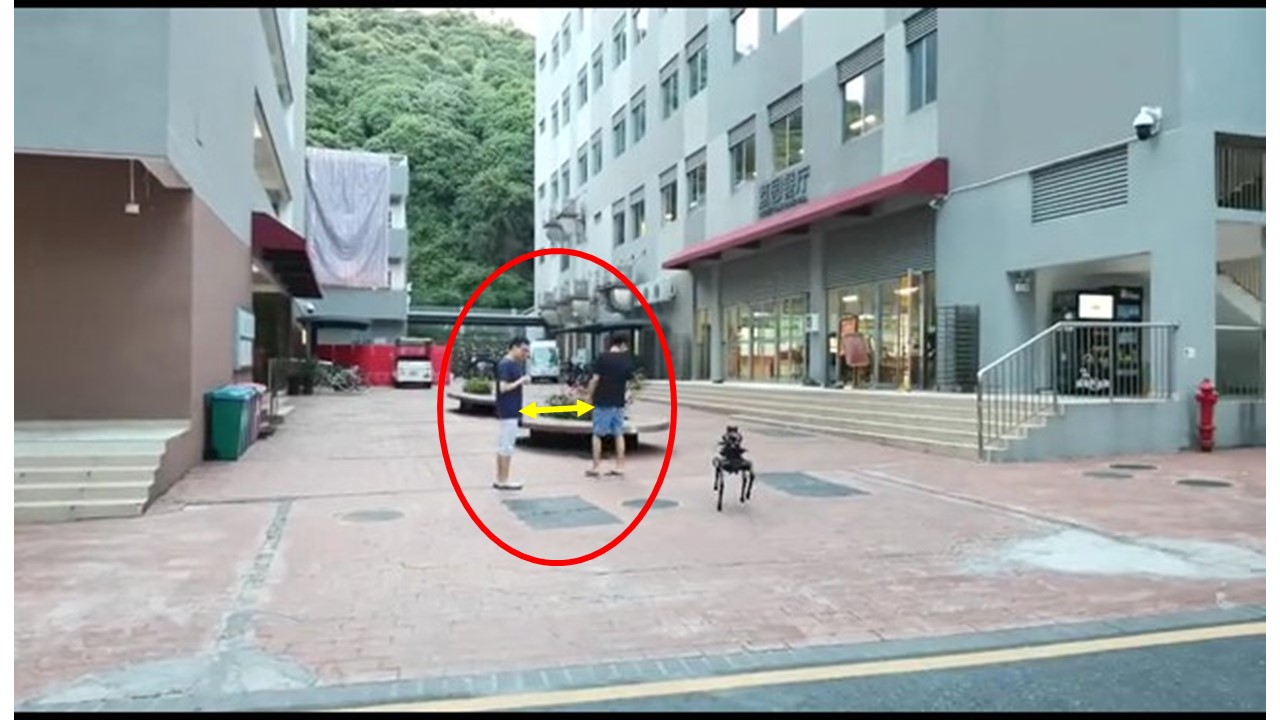}
    \end{subfigure} 
    \vskip\baselineskip
    \begin{subfigure}{0.49\linewidth}
    \includegraphics[trim={10 10 10 10},clip, width=\textwidth]{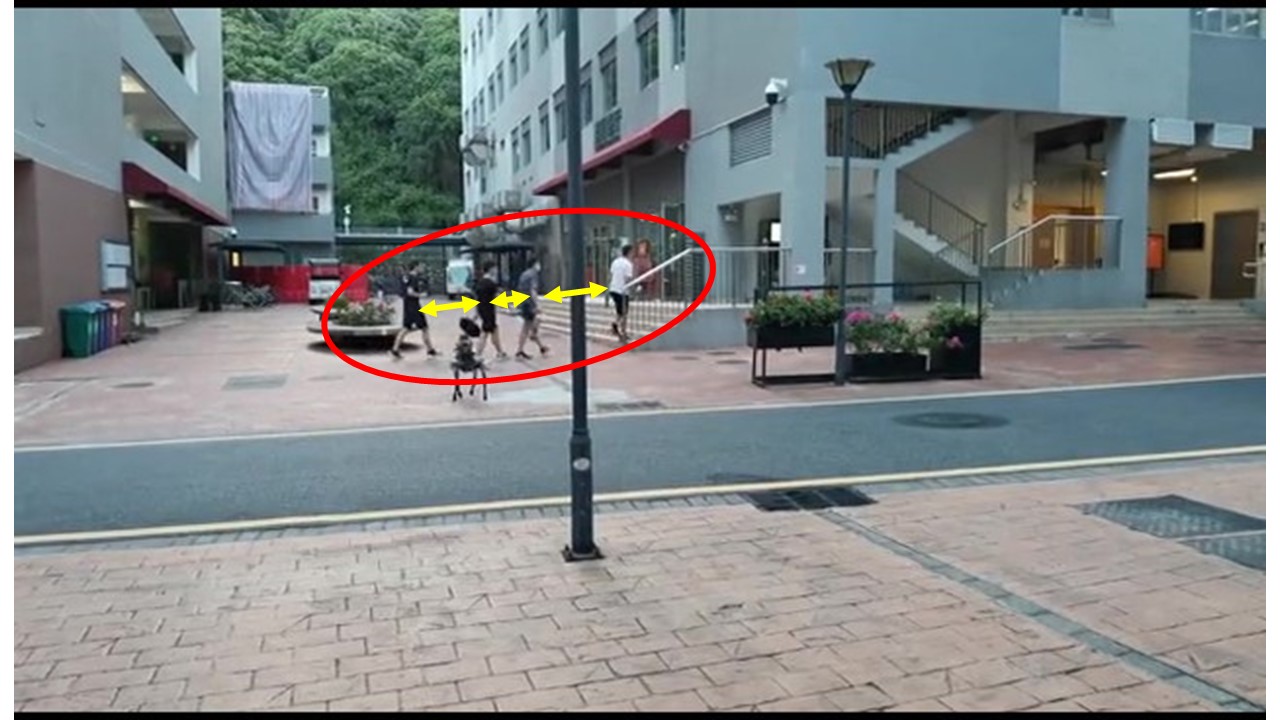}
    \end{subfigure}
    \begin{subfigure}{0.49\linewidth}
    \includegraphics[trim={10 10 10 10},clip,width=\textwidth]{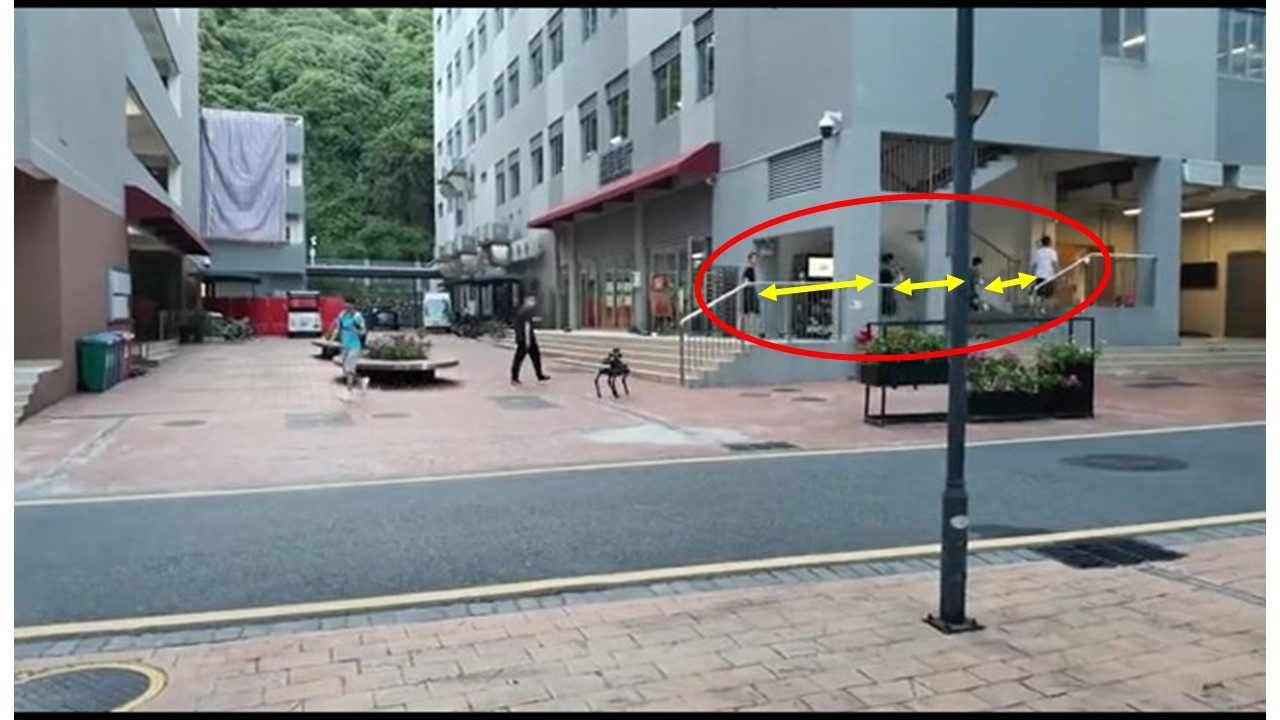}
    \end{subfigure}
    \caption{Examples from the real-world experiment. The top and bottom images describes two different scenarios. Left: The robot detected and approached the crowds, then persuaded them to keep social distance. Right: The crowds density decreased.}

    \label{fig:real_test_example}
\end{figure*}

Finally, we integrate all the above modules together, and investigate whether the robot can navigate in the complex urban environments with satisfiable social distancing effectiveness without terrifying general citizens. The real-world experiment was conducted in two public areas including a university campus and a park. 
Figure~\ref{fig:real_test_example} shows some examples from the real world experiment.

The result shows that our robot successfully fulfills the task of social distancing. For people who have been interacted with our robot, about half of them followed the robot suggestions. For the other people, most of them glanced at the robot and then just walked away, some of them stopped and looked at the robot. It's worth to notice that at the time of our experiment, there were no existing COVID-19 patients in the testing city, which tends to reduce the pedestrian's compliance with the verbal social distancing commands. During the experiment, we selected some people randomly, then asked them about their attitude towards the robot and why they followed/didn't follow the robot's advice. Some people reported that they felt it is a great idea to use the surveillance robot and they thought the robot's advice is reasonable. Besides, many people reported the robot looks like it came from the world of science fiction so they were very curious about the robot. However, some people felt the robot is not friendly enough so they just wanted to walk away. For the people who ignored the robot's advice, most of them said that the pandemic is not severe so they felt it's unnecessary to keep the distance.

\section{Conclusion}
\label{sec:conclusion}

In the context of the COVID-19 pandemic, we develop the autonomous surveillance robot system to promote social distancing. The robot system is mainly composed of social distance detection, urban navigation, and intelligent voice interaction. The legged robot shows good adaptation to different terrain so that they can work well in human life scenarios. The real-world experiment also demonstrates our robot successfully keeps human's social distance. In this end, we successfully deploy the system in a real environment to prevent the spread of COVID-19.  

{
\bibliographystyle{IEEEtran}
\bibliography{references}
}

\end{document}